\documentclass[letterpaper,journal]{IEEEtran}
\usepackage{amsmath,amsfonts}
\usepackage{algorithmic}
\usepackage{array}
\usepackage[caption=false,font=normalsize,labelfont=sf,textfont=sf]{subfig}
\usepackage{hyperref}
\usepackage{orcidlink}
\usepackage{textcomp}
\usepackage{stfloats}
\usepackage{url}
\usepackage{verbatim}
\usepackage{graphicx}
\usepackage{cite}
\usepackage[ruled,vlined]{algorithm2e}
\usepackage[table]{xcolor} 
\usepackage{booktabs}   
\usepackage{multirow}   
\usepackage{float}      
\SetKwInOut{Input}{Input}
\SetKwInOut{Output}{Output}
\begin{document}
\title{AccidentSim: Generating Vehicle Collision Videos with Physically Realistic Collision Trajectories from Real-World Accident Reports}
\author{
    Xiangwen Zhang~\orcidlink{0009-0000-2717-9838}, 
    Qian Zhang~\orcidlink{0009-0007-4824-3158}, 
    Longfei Han~\orcidlink{0000-0003-2135-6228}, \\
    Qiang Qu~\orcidlink{0000-0002-6648-5050}, 
    Xiaoming Chen~\orcidlink{0000-0002-7503-3021}, 
    and Weidong Cai~\orcidlink{0000-0003-3706-8896}
    \thanks{Corresponding author: Xiaoming Chen.}
    
    \thanks{Xiangwen Zhang, Qian Zhang, Longfei Han, and Xiaoming Chen are with the School of Computer and Artificial Intelligence, Beijing Technology and Business University, Beijing 100048, China (e-mail: xiangwenzhang223@163.com; qian.zhang@st.btbu.edu.cn; draflyhan@gmail.com; xiaoming.chen@btbu.edu.cn).}
    
    \thanks{Qiang Qu and Weidong Cai are with The University of Sydney, Sydney, NSW 2006, Australia (e-mail: vincent.qu@sydney.edu.au; tom.cai@sydney.edu.au).}
}
\markboth{Journal of \LaTeX\ Class Files,~Vol.~14, No.~8, August~2021}%
{Shell \MakeLowercase{\textit{et al.}}: A Sample Article Using IEEEtran.cls for IEEE Journals}
\maketitle

\begin{abstract}
Collecting real-world vehicle accident videos for autonomous driving research is challenging due to their rarity and complexity. While existing video generation methods can produce visually realistic content, they often lack physical realism and fail to generate accurate post-collision vehicle trajectories. In this paper, we introduce AccidentSim, a framework designed to generate physically realistic vehicle collision videos. Specifically, AccidentSim extracts and utilizes physical cues and contextual information from real-world accident reports. It then uses a physical simulator to replicate post-collision vehicle trajectories based on the information from these reports, creating a collision trajectory dataset. This dataset is subsequently used to fine-tune a language model, AccidentLLM, enabling it to predict novel, physically consistent post-collision trajectories across various driving scenarios based on user-provided accident descriptions. Finally, we synthesize collision videos by compositing vehicles that follow the predicted trajectories onto rendered backgrounds. Experiments show that AccidentSim obtains the best overall physical-realism and semantic-consistency results among the compared methods and that its generated scenarios reduce the average collision rate of the evaluated driving policy.
\end{abstract}


\begin{figure*}[hbt!]
\vspace{-0.5cm}
\centering
\includegraphics[width=\textwidth]
{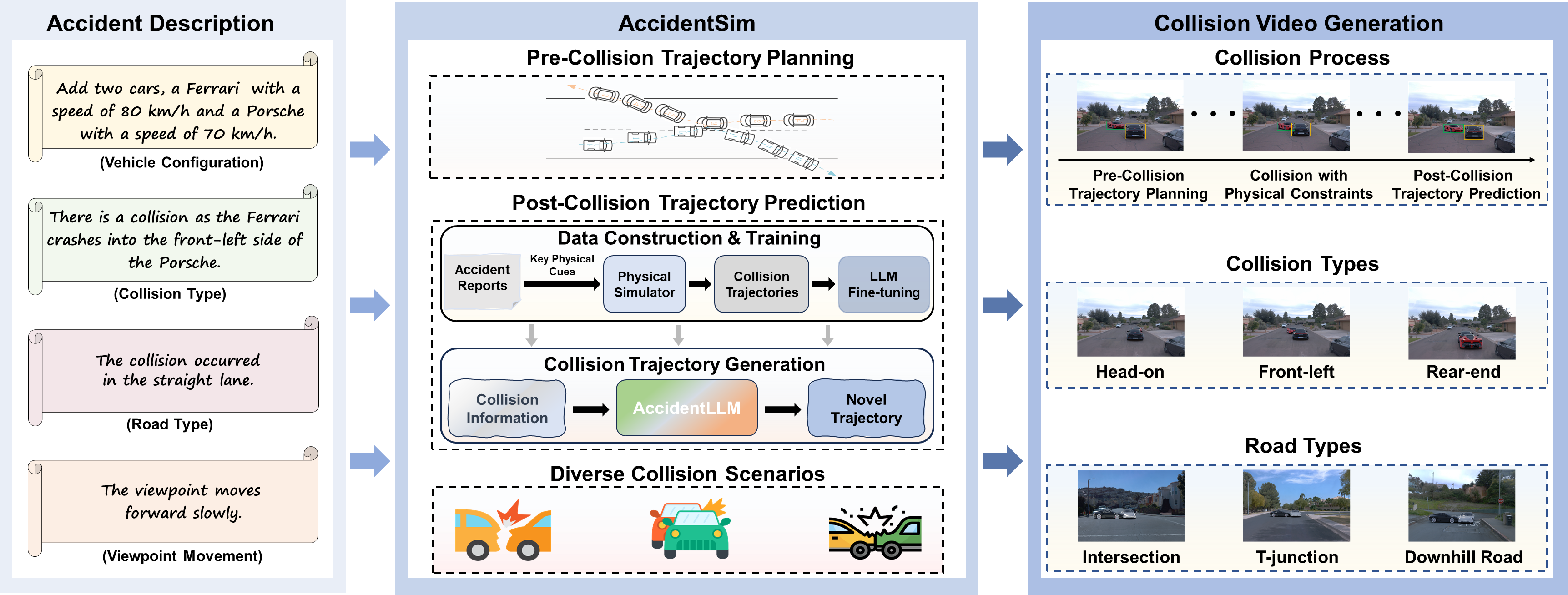}
\caption{
AccidentSim generates vehicle collision videos from user-provided accident descriptions, producing physically realistic collision trajectories of vehicles across diverse collision and road types, all according to the user-specified viewpoints. Particularly, AccidentSim extracts physical cues and contextual information from real-world accident reports and feeds them into a physical simulator to generate physically constrained collision trajectories. These trajectories are then used to fine-tune a language model, termed AccidentLLM, enabling it to predict novel, physically consistent trajectories directly from the collision information extracted from the user-provided accident descriptions, extending beyond the collision scenarios contained in the original accident reports.
}
\label{fig:trigger} 
\end{figure*}
\section{Introduction}
\IEEEPARstart{R}{esearch} and development in autonomous driving critically depend on access to diverse and comprehensive training data to ensure robust performance across various driving scenarios. However, existing datasets, such as KITTI \cite{geiger2013vision}, Cityscapes \cite{cordts2016cityscapes}, nuScenes \cite{caesar2020nuscenes}, Waymo \cite{sun2020scalability}, and BDD100K \cite{yu2020bdd100k}, primarily capture normal driving conditions and contain relatively few safety-critical events such as vehicle collisions. This data imbalance arises from the inherent difficulty and prohibitive cost of collecting rare but safety-critical scenarios in real-world environments. This phenomenon is known as the Curse of Rarity (CoR) \cite{liu2024curse}, which refers to the low occurrence rate of safety-critical events in high-dimensional spaces. In autonomous driving systems, CoR limits the ability of deep learning models to learn effective responses to these rare but crucial situations, ultimately affecting the reliability and safety of such systems. 

With recent advances in video generation models, synthesizing vehicle collision videos has become a feasible, low-cost, and promising alternative to expensive real-world data collection. Such generated videos, if \textit{physically realistic}, could augment existing datasets and enhance the training of autonomous driving systems, ultimately improving reliability and safety in critical scenarios. However, current video generation methods, including state-of-the-art models such as Kling~\cite{klingai2025}, Gen-4~\cite{Runway2025}, and Jimeng~\cite{Jimengai2025}, as well as specialized driving-scenario generators like DriveDreamer~\cite{wang2024driving}, Panacea~\cite{wen2024panacea}, Cosmos~\cite{agarwal2025cosmos}, and HoloDrive~\cite{wu2024holodrive}, still struggle to produce physically realistic vehicle collisions. In particular, these approaches often fail to reproduce physically realistic collision trajectories that conform to real-world physical constraints and dynamics. This limitation arises because these video generation methods are not explicitly designed to incorporate essential physical cues and constraints. For instance, existing models typically overlook key physical factors, such as vehicle momentum, impulse, and mass distribution, that are critical for accurately modeling real-world collision behaviors. Consequently, while the videos generated by these models may be visually realistic, they lack physical realism, failing to capture the true underlying dynamics of vehicle collisions.

To address these challenges, we propose AccidentSim, a framework that introduces a novel approach to generating physically realistic vehicle collision videos based on \textit{real-world accident reports}. For instance, accident reports from the National Motor Vehicle Crash Causation Survey (NMVCCS)~\cite{nhtsa-report} in the United States document actual vehicle collisions across diverse vehicle models, collision types, road types, and traffic conditions. AccidentSim leverages the rich physical cues and contextual information embedded in such reports to synthesize vehicle-collision videos with physically realistic post-collision trajectories.
As illustrated in Figure~\ref{fig:trigger}, AccidentSim first extracts key physical cues, such as vehicle speed, collision angle, and mass, together with contextual information, including road type and collision type, from the accident reports. These cues serve as the physical foundation for constructing realistic collision simulations. The extracted physical cues are then fed into a physical simulator, such as CARLA~\cite{Dosovitskiy17}, which enforces corresponding physical constraints to ensure that the simulated post-collision vehicle trajectories adhere to real-world dynamics. However, physical simulators have inherent limitations. For example, physical simulators are typically computationally expensive and require fully specified physical parameters, making them unable to directly respond to user-provided accident descriptions for generating diverse post-collision trajectories. To overcome these challenges, we first construct a dataset from the collision trajectories generated by the simulator using real-world accident reports. This dataset is then used to fine-tune a large language model, termed AccidentLLM, within the AccidentSim framework. As a result, AccidentLLM can generate novel, physically realistic collision trajectories directly from the collision information extracted from the user-provided accident descriptions, extending beyond the scenarios captured in existing accident reports. In this way, AccidentLLM effectively bypasses the expensive simulation process through physics distillation, learning to translate under-specified semantic inputs into physically realistic trajectories efficiently and realistically. Finally, AccidentSim synthesizes collision videos by compositing the predicted trajectories onto vehicles with rendered backgrounds, according to user-specified viewpoints.

In summary, AccidentSim integrates physically grounded cues with physical constraints to generate videos representing diverse vehicle-collision scenarios. By connecting accident-report evidence with controllable simulation, AccidentSim provides an approach to augmenting scarce collision data for training and evaluating autonomous driving systems. 

Our contributions are three-fold:
\begin{itemize}
\item We propose AccidentSim, a framework that extracts and leverages physical cues and contextual information from real-world accident reports, to generate vehicle collision videos with physical simulation, ensuring physically realistic post-collision trajectories.
\item We develop AccidentLLM, a language-driven module fine-tuned on simulated collision trajectories for post-collision prediction. Combined with our pre-collision planner, it generates complete, physically realistic trajectories from user-provided accident descriptions.
\item We evaluate AccidentSim across multiple road types, collision types, and driving scenarios, including its utility for downstream trajectory prediction and safety-oriented policy training. 
\end{itemize}

\section{Related Work}

\subsection{Safety-Critical Scenarios and Driving Datasets}
The foundation of autonomous driving research relies on large-scale datasets, such as KITTI \cite{geiger2013vision}, Cityscapes \cite{cordts2016cityscapes}, nuScenes \cite{caesar2020nuscenes}, Waymo Open Dataset \cite{sun2020scalability}, and BDD100K \cite{yu2020bdd100k}. While providing many normal-driving trajectories, they contain relatively few safety-critical vehicle collisions. To address this, early efforts constructed specialized accident datasets, such as CADP \cite{8639160}, ISSAFE \cite{9636109}, and CCD \cite{BaoMM2020}, which are enriched with spatiotemporal and environmental annotations. Recently, vision-based traffic accident anticipation \cite{fang2023vision} has tackled collisions as challenging out-of-distribution (OOD) events. Modern predictive frameworks enhance early detection and fine-grained analysis through multimodal feature alignment \cite{patera2025multi}, multilevel knowledge distillation \cite{yu2023multiple}, and single-stage text-prompt adaptation \cite{liang2024text}. Furthermore, EQ-TAA \cite{fang2025eq} employs video diffusion to synthesize pseudo-accident pairs to mitigate data bias. Real-world accident datasets nevertheless remain constrained by acquisition cost, limited diversity, and often fixed viewpoints, while anticipation methods analyze observed events rather than generate controllable crash scenarios. To reduce dependence on real-world data collection, adversarial scenario generation and trajectory optimization have been widely explored. Methods like AdvSim \cite{Wang2021AdvSimGS}, Learning-to-Collide \cite{Ding2020LearningTC}, and CAT \cite{zhang2023cat} employ adversarial training for challenging pre-collision scenarios, while STRIVE \cite{rempe2022generating} and others \cite{zhang2022adversarial} optimize latent traffic models evaluated in platforms like SafeBench \cite{xu2022safebench}. However, these approaches primarily target pre-collision interactions and obstacle avoidance rather than physical impact and post-collision dynamics.

\subsection{Trajectory Prediction and Language-Guided Generation}
Accurate vehicle motion forecasting is core to driving simulation. Traditional multi-agent trajectory prediction models \cite{shi2022motion, jia2023hdgt, nayakanti2023wayformer, feng2024unitraj} rely heavily on historical states, latent sequence modeling \cite{girgis2021latent}, and high-definition maps. Recently, language-guided traffic generation driven by large language models (LLMs) \cite{dubey2024llama} has emerged. Frameworks like LCTGen \cite{tan2023language}, CTG++ \cite{zhong2023language}, TrafficGen \cite{feng2022trafficgen}, and Traj-LLM \cite{yang2025trajectory} control traffic simulation via natural language. MotionLM \cite{seff2023motionlm} formulates multi-agent forecasting as language modeling, whereas MotionCLIP \cite{tevet2022motionclip} maps motion generation into a joint language--motion representation. To bridge high-level reasoning and low-level control, recent studies distill multi-modal LLM knowledge into efficient vision-based planners \cite{hegde2025distilling}, enhancing generalizability in critical scenarios without prohibitive inference costs. Alternatively, to manage online querying overhead, AdaDrive \cite{zhang2025adadrive} proposes an adaptive collaborative system. By dynamically invoking the LLM exclusively during complex scenarios, it balances high-level reasoning with fast conventional planners.
\subsection{Physical Simulation and Collision Modeling}
Early analytical models \cite{pawlus2013data, zhou2008collision} utilized momentum, restitution, and friction to predict post-impact motion. Modern research integrates these principles into 3D simulators like CARLA \cite{Dosovitskiy17}, LGSVL \cite{rong2020lgsvl}, and UniSim \cite{yang2023unisim} to model vehicle dynamics under physical constraints. While these engines provide physics-based simulation, integrating them into data-driven generative pipelines remains challenging. Recently, SoVAR \cite{guo2024sovar} and AC3R \cite{huynh2019ac3r} parsed real-world police reports from NMVCCS \cite{nhtsa-report} to reconstruct accidents in simulators. Concurrently, hierarchical models \cite{gao2025diversifying} decompose scenes into reusable assets and perturbed trajectories, utilizing Graph Conditional VAEs to instantiate challenging scenarios in CARLA. However, physical simulation can be computationally expensive, requires detailed parameter specification, and cannot directly interpret underspecified prompts. AccidentSim instead uses the simulator to construct a trajectory dataset and distills this data into AccidentLLM, avoiding online physical simulation during trajectory prediction.

\subsection{Controllable Video Generation and World Models}
The rapid evolution of diffusion models has profoundly transformed autonomous driving simulation, shifting the research focus toward creating immersive and controllable visual environments. Foundation generative models such as Kling 2.0 \cite{klingai2025}, Gen-4 \cite{Runway2025}, Jimeng 3.0 \cite{Jimengai2025}, and CogVideoX \cite{yang2024cogvideox} have established new benchmarks in photorealism. Tailored for the driving domain, generative world models including DriveDreamer \cite{wang2024driving}, DriveDiffusion \cite{li2023drivingdiffusion}, MagicDrive \cite{gao2023magicdrive}, Panacea \cite{wen2024panacea}, HoloDrive \cite{wu2024holodrive}, Dreamforge \cite{mei2024dreamforge}, and Cosmos \cite{agarwal2025cosmos} successfully synthesize normal driving videos conditioned on layouts, text, or 3D geometry. Concurrently, advancements in NeRF and 3D Gaussian Splatting (3DGS), such as S-NeRF \cite{S-NeRF}, ChatScene \cite{zhang2024chatscene}, NeuroNCAP \cite{ljungbergh2024neuroncap}, Editable Scene Simulation \cite{wei2024editable}, DrivingGaussian \cite{zhou2024drivinggaussian}, and Street Gaussians \cite{yan2024street}, enable editable large-scale dynamic scene simulation and novel view synthesis. To mitigate severe class imbalances between rare road objects and dominant background layouts, diffusion-guided BEV map generators \cite{gao2025boundary} incorporate boundary-aware losses, increasing minority class weights for accurate scene synthesis.

Moving beyond normal conditions, specialized frameworks like Challenger \cite{xu2025challenger} provide affordable adversarial video generation. Utilizing physics-aware trajectory refinement and tailored scoring, they synthesize aggressive scenarios like cut-ins to stress-test autonomous systems. To overcome the prohibitive computational overhead of high-resolution multi-view synthesis, DriveScape \cite{wu2025drivescape} efficiently fuses multi-view features with 3D road structures. Furthermore, MagicDrive-V2 \cite{gao2025magicdrive} integrates spatial-temporal conditional encoding into Diffusion Transformers, achieving unprecedented high-resolution, long video generation with precise geometric and textual control.

\begin{figure*}[ht]
    \centering
    \includegraphics[width=\textwidth]
    {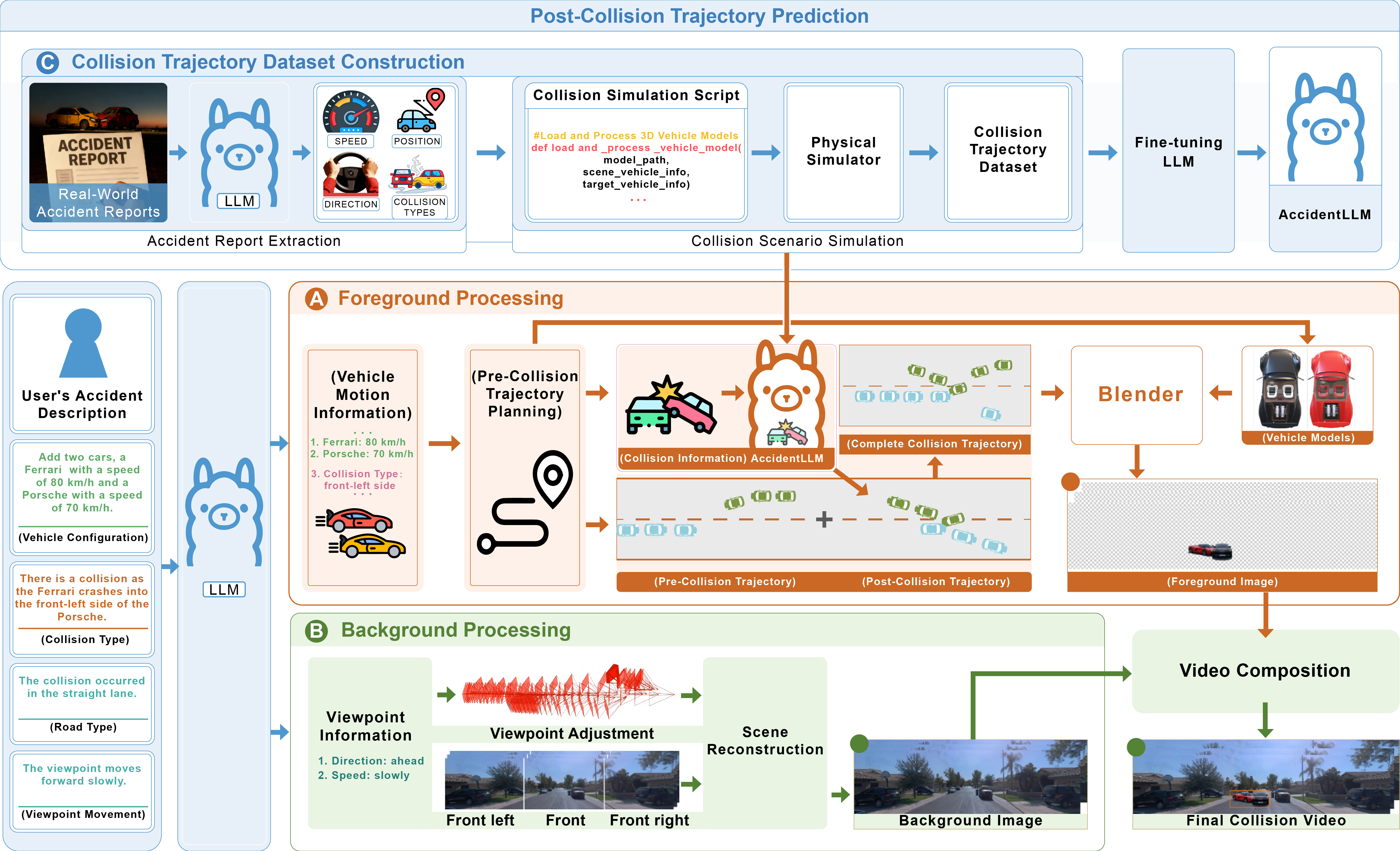}
    \captionsetup{width=\textwidth}  
    \caption{Architecture of AccidentSim: AccidentSim extracts physical cues, such as vehicle speeds and collision types, along with contextual information like road types, from accident reports. This extracted information is then fed into a physical simulator, such as CARLA, to generate collision trajectories, which form a dataset. The dataset is subsequently used to fine-tune a language model, AccidentLLM (based on the Llama model), enabling it to adapt to various scenarios from user-provided accident descriptions and generate novel trajectories. Building on this foundation, AccidentSim leverages user accident descriptions to plan pre-collision trajectories, generating both the pre-collision trajectory and relevant collision information, such as vehicle speeds, collision angles, and more. The collision information is then processed by AccidentLLM, which produces novel, physically realistic post-collision trajectories, resulting in complete collision trajectories. These realistic trajectories are composited onto vehicles to generate the foreground, which is then combined with rendered backgrounds to produce collision videos across various scenarios.
}
    \label{fig:framework}
\end{figure*}
\section{Methodology}
AccidentSim first utilizes a large language model to extract key vehicle motion information (such as vehicle models, collision types, and intended video viewpoint movements) from the user-provided accident description. This extracted information is then processed by two core modules: the \textbf{Pre-Collision Trajectory Planning} module, which generates the trajectories leading to the collision, and the \textbf{Post-Collision Prediction} module, which generates the trajectories after the collision. Together, these modules produce a complete vehicle collision trajectory. Subsequently, in the following \textbf{Collision Video Generation} phase, the \textbf{Foreground Processing} module integrates the generated collision trajectory with user-specified vehicle models to produce foreground images. In the \textbf{Background Processing} module, background images are synthesized based on the user-specified viewpoint movement. Finally, the \textbf{Video Composition} module merges the foreground and background images to generate a vehicle collision video.
\subsection{Pre-Collision Trajectory Planning} \label{sec:pre_collision}
At this stage, AccidentSim produces pre-collision trajectories leading up to the collision, along with associated information such as vehicle speed. Algorithm~\ref{alg:trajectory} summarizes three phases. First, in the \textbf{Initial State Extraction} phase, essential vehicle and collision parameters are extracted from the user-provided accident description to establish the initial state for each vehicle and define the collision target. Second, in the \textbf{Lane and Path Selection} phase, the algorithm identifies suitable lanes based on vehicle dynamics and the scene map, forming potential lane combinations to ensure feasible paths for collision trajectory planning. Third, in the \textbf{Trajectory Generation and Validation} phase, valid pre-collision trajectories are generated using the selected lane combinations and evaluated against collision criteria to yield a final set of feasible pre-collision trajectories. 
\begin{algorithm}[t]
\caption{Pre-Collision Trajectory Planning}
\label{alg:trajectory}
\Input{Vehicle motion information $\mathcal{I}$; scene map $\mathcal{M}$ containing lane nodes}
\Output{Valid pre-collision trajectories $\mathcal{T}$; imminent-collision state $\mathcal{S}$}
\SetKwBlock{PhaseOne}{\textbf{Phase 1: Initial State Extraction}}{}
\PhaseOne{
    $\{\mathcal{C}, n, D, V, A, \theta\} \gets \textsc{ExtractState}(\mathcal{I})$\;
    $P_0 \gets \textsc{DetermineStartingPoints}(D, \theta)$\;
}
\SetKwBlock{PhaseTwo}{\textbf{Phase 2: Lane and Path Selection}}{}
\PhaseTwo{
    $L \gets \textsc{ParseMap}(\mathcal{M})$; $\mathcal{L} \gets \emptyset$\;
    \ForEach{lane $l_i \in L$}{
        \If{\textsc{IsValidLane}($l_i$, $\theta$, $\mathcal{C}$)}{
            $\mathcal{L} \gets \mathcal{L} \cup \{l_i\}$\;
        }
    }
    $\mathcal{G} \gets \textsc{GatherLaneCombinations}(\mathcal{L},n,D,V,A,\theta)$\;
}
\SetKwBlock{PhaseThree}{\textbf{Phase 3: Trajectory Generation and Validation}}{}
\PhaseThree{
    \ForEach{lane combination $G_i \in \mathcal{G}$}{
        \If{\textsc{MeetsCollisionCriteria}($G_i$, $\mathcal{C}$)}{
            $G_s \gets G_i$; \textbf{break}\;
        }
    }
    $\mathcal{T}_{raw} \gets \textsc{GenerateTrajectory}(G_s, P_0)$\;
    $\mathcal{T} \gets \textsc{ValidateTrajectories}(\mathcal{T}_{raw})$\;
    $\mathcal{S} \gets \textsc{GetImminentCollisionState}(\mathcal{T}, V, A)$\;
}
\Return{$\{\mathcal{T},\mathcal{S}\}$}
\end{algorithm}
\subsubsection{Initial State Extraction}
In this phase, the proposed algorithm initiates by extracting essential vehicle and collision parameters from the vehicle motion information $\mathcal{I}$, which includes the dynamics of all vehicles involved in the scenario. Using this data, the algorithm isolates variables such as the collision point $\mathcal{C}$, vehicle count $n$, distances $D$, velocities $V$, accelerations $A$, and orientation angles $\theta$. This extraction process establishes a comprehensive view of each vehicle's current state and potential collision zones. The algorithm then determines the starting points $P_0$ for each vehicle by factoring in directional and dynamic properties.

\subsubsection{Lane and Path Selection}
This phase focuses on selecting suitable lanes within the scene map $\mathcal{M}$ to enable feasible trajectory planning. The proposed algorithm begins by parsing the scene map to identify available lanes $L$, ensuring that each candidate lane is assessed for compatibility with the vehicle’s current orientation $\theta$ and collision constraints $\mathcal{C}$. A subset of valid lanes $\mathcal{L}$ is then assembled, containing only those that meet the criteria defined by vehicle orientation and other relevant dynamics. After identifying valid lanes, the algorithm generates various lane combinations $\mathcal{G}$ by analyzing lane connectivity and spatial feasibility in relation to the parameters $n$, $D$, $V$, $A$, and $\theta$.

\subsubsection{Trajectory Generation and Validation} This phase involves generating and validating trajectories based on the preselected lane combinations. For each lane combination $G_i$ in $\mathcal{G}$, the proposed algorithm checks whether it meets specific collision criteria using the collision parameters $\mathcal{C}$. Upon identifying a valid lane combination $G_s$ that satisfies these conditions, it proceeds to generate the trajectory paths $\mathcal{T}$. This process incorporates the starting points $P_0$ and valid lane combination $G_s$ to form trajectories that are both dynamically feasible and collision-aware. Each trajectory is then evaluated to ensure it conforms to the criteria established for pre-collision scenarios. The final output consists of the validated set of trajectories $\mathcal{T}$ and the associated imminent-collision information $\mathcal{S}$. AccidentLLM then uses $\mathcal{S}$ to predict the post-collision trajectories described in Section~\ref{sec:post_collision}.

\subsection{Post-Collision Trajectory Prediction} \label{sec:post_collision}
To address the challenge of collecting real-world post-collision trajectories, we leverage physical cues extracted from accident reports to train a customized version of the Llama model \cite{dubey2024llama}. The resulting model, termed \textit{AccidentLLM}, is capable of responding to the collision information extracted from the user-provided accident descriptions, and predicting novel post-collision trajectories that are adaptable to various road conditions. 

As shown in Figure \ref{fig:framework} (c), the proposed prediction is structured into three stages: (1) \textbf{Accident Report Extraction}, where collision-specific information is extracted from accident reports; (2) \textbf{Collision Scenario Simulation}, which generates physics-constrained post-collision trajectories; and (3) \textbf{Fine-tuning AccidentLLM}, where the extracted attributes and simulated scenarios are used to tailor the model for accurate trajectory prediction.

\subsubsection{Accident Report Extraction}\label{Accident_Report_Extraction}
\begin{figure*}[t]
    \centering
    \includegraphics[width=0.78\textwidth]{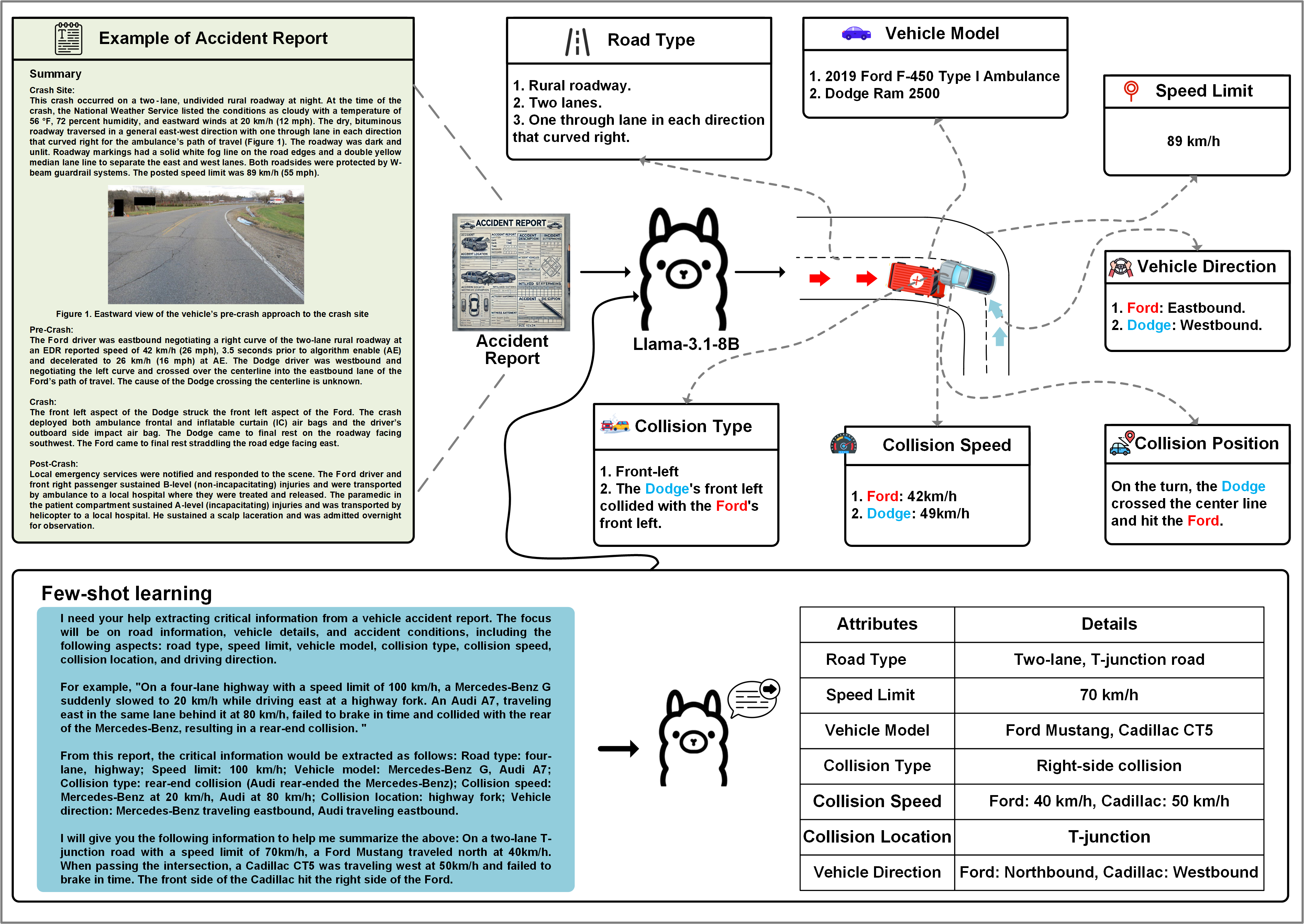}
    \caption{Structured extraction of contextual and physical cues from an accident report. A few-shot Llama-3.1-8B extractor converts free-form narratives into road, vehicle, and collision attributes used to initialize simulation and trajectory generation.}
    \label{fig:accident_report}
\end{figure*}

The real-world accident reports used in our work, sourced from NMVCCS \cite{nhtsa-report}, provide critical data for supporting collision simulations, encompassing pre-accident environmental conditions, collision dynamics, and post-accident outcomes. Prior work, such as SoVAR \cite{guo2024sovar}, employs a template-based approach combined with language models to extract key information from accident reports. However, due to the inherent variability and complexity of natural language descriptions in these reports, a single-template strategy lacks the flexibility to adapt to diverse accident scenarios. To overcome this limitation, we employ a Llama model \cite{dubey2024llama} with a few-shot learning strategy to automatically extract essential information from accident reports, including pre-accident environmental conditions, road features, and driver behaviors; in-collision details such as vehicle positioning, speed, and collision type; and post-accident information including emergency responses. 
Figure~\ref{fig:accident_report} illustrates this structured extraction process. The resulting attribute record specifies the road topology, speed limit, vehicle models, collision configuration, impact speeds, collision location, and travel directions. These attributes provide a common interface between an under-specified natural-language report and the deterministic initialization required by the trajectory planner and CARLA simulator.

\subsubsection{Collision Scenario Simulation}\label{sec:collision_sim}
In the accident simulation phase, we develop a scenario reconstruction pipeline to instantiate report-derived collision scenarios and construct a physics-constrained collision dataset for the subsequent fine-tuning of AccidentLLM. The process begins by partitioning the extracted accident-report information into vehicle-specific attributes, such as velocity and location, and road-related attributes. Using RoadRunner \cite{RoadRunner}, we construct 3D road models and integrate them into the CARLA simulation environment. CARLA's vehicle dynamics, together with the extracted motion attributes and configured physical parameters, provide physics-based motion and collision responses. 
Specifically, road-surface properties determine the friction coefficients, while road slope and gravity influence vehicle acceleration and deceleration. As each vehicle approaches the designated collision point, its speed, position, and orientation at impact are recorded together with the resulting post-collision trajectory. These simulated trajectories provide supervision for fine-tuning the Llama model \cite{dubey2024llama} to predict post-collision motion across the reconstructed scenarios. 

\subsubsection{Fine-tuning AccidentLLM} \label{sec:accidentllm}
After obtaining the collision dataset from the previous process, we fine-tune the Llama model \cite{dubey2024llama} via LoRA \cite{hu2021lora} to specialize in predicting novel post-collision trajectories, resulting in a model we term AccidentLLM. This supervised fine-tuning maps imminent-collision states derived from user-provided descriptions to post-collision trajectories.

To optimize AccidentLLM for precise trajectory prediction, we apply supervised fine-tuning using an $L_1$ loss function that measures the absolute discrepancy between predicted and ground truth trajectories across both spatial and rotational dimensions. The trajectory matching loss is defined as:
\begin{equation}
L_{\text{traj}} = 
\frac{1}{N} \sum_{i=1}^{N} 
\left(
\|\hat{\mathbf{p}}_i - \mathbf{p}_i^*\|_1
+ 
\|\hat{\boldsymbol{\theta}}_i - \boldsymbol{\theta}_i^*\|_1
\right)
\end{equation}

\noindent where $\hat{\mathbf{p}}_i = [\hat{p}_{x,i}, \hat{p}_{y,i}, \hat{p}_{z,i}]^\top$ and $\mathbf{p}_i^*$ represent the predicted and ground-truth positions at time step $i$, respectively, while $\hat{\boldsymbol{\theta}}_i$ and $\boldsymbol{\theta}_i^*$ denote the predicted and true rotational vectors. $N$ is the total number of time steps in the trajectory sequence. This formulation ensures that both translational and rotational components are jointly optimized, allowing AccidentLLM to capture complex post-collision dynamics and produce physically consistent trajectory predictions in diverse crash scenarios.

\subsection{Collision Video Generation}
\subsubsection{Foreground Processing}
As shown in Figure \ref{fig:framework} (a), the vehicle motion information extracted from the user-provided accident description undergoes \textbf{Pre-Collision Trajectory Planning}, which identifies the involved vehicle models and generates both the pre-collision trajectories and key collision parameters, such as vehicle speed and collision type (e.g., rear-end collision). These results are then passed to \textbf{AccidentLLM}, which infers physically realistic post-collision trajectories to complete the full motion paths of the vehicles involved. Based on these complete trajectories, foreground images of the physically realistic vehicle collision are rendered.

\subsubsection{Background Processing}\label{sec:background}
As shown in Figure \ref{fig:framework} (b), the background processing consists of two phases: viewpoint adjustment and background rendering. In the viewpoint adjustment phase, we first parse the intended video viewpoint movements, extracted from the user-provided accident description, into positional and angular shift parameters corresponding to the target viewpoint. These shift parameters are then converted into the appropriate extrinsic transformation matrix and combined with the initial parameters to yield an updated viewpoint configuration. In the background rendering phase, if a novel viewpoint needs to be rendered, we apply the scene reconstruction method from ChatSim \cite{wei2024editable}. Otherwise, the original frame from the Waymo Open Dataset is used as the background. Specifically, ChatSim’s scene reconstruction method incorporates multi-camera alignment and luminance consistency rendering techniques, ensuring that the generated background seamlessly and naturally aligns with the vehicle’s viewpoint.

\subsubsection{Video Composition}\label{sec:composition}
In the final video composition stage, we use alpha-channel compositing to combine the rendered foreground vehicles with the background frames while preserving the physically grounded trajectories generated in the preceding stages.

\section{Experiments}
To evaluate AccidentSim, we use the Waymo Open Dataset and NHTSA accident reports (Section~\ref{sec:datasets}) and organize the experiments around five claims. First, Section~\ref{sec:extraction_eval} evaluates report parsing and executable scenario reconstruction. Second, Section~\ref{sec:traj_quality} measures post-collision accuracy, physical consistency, and text alignment. Third, Section~\ref{sec:prediction_enhancement} tests distribution-level fidelity and downstream trajectory-prediction utility. Fourth, Section~\ref{sec:video_quality} evaluates visual quality and physical realism. Finally, Section~\ref{sec:collision_reduction} tests whether the generated scenarios improve an autonomous-driving policy.

\subsection{Datasets}\label{sec:datasets}
We conducted experiments on the Waymo Open Dataset~\cite{sun2020scalability} and real-world accident reports maintained by the NHTSA. The Waymo Open Dataset contains data from 1,150 driving scenarios, which is suitable for covering the diverse scenarios required by AccidentSim. Each scenario in Waymo is recorded for 20 seconds and includes synchronized data from five LiDAR sensors and five high-resolution pinhole cameras. The NHTSA accident reports encompass a wide range of vehicle models, collision types, road types, and traffic conditions, making them suitable for use with AccidentSim.
\subsection{Accident-Report Extraction and Reconstruction}\label{sec:extraction_eval}
We first evaluate the two stages that convert accident narratives into executable collision scenarios. For information extraction, we compare AccidentSim with SoVAR~\cite{guo2024sovar}, its pattern-free variant SoVAR\_N, and the rule-based AC3R~\cite{huynh2019ac3r}. Table~\ref{tab:extract_accuracy} reports exact-match accuracy for six attributes spanning road context and dynamic objects. AccidentSim performs best on five attributes and remains competitive on crash type, showing that the flexible few-shot extractor reliably supplies the parameters required by subsequent simulation.
\begin{table*}[t]
    \centering
    \caption{Information-extraction accuracy (\%) from accident reports. SoVAR\_N denotes SoVAR without linguistic-pattern guidance. Best values are shown in bold.}
    \label{tab:extract_accuracy}
    \resizebox{0.82\textwidth}{!}{
    \begin{tabular}{l c c c c c c}
        \toprule
        \multirow{2}{*}{\textbf{Method}} & \multicolumn{3}{c}{\textbf{Road Context}} & \multicolumn{3}{c}{\textbf{Dynamic Objects}} \\
        \cmidrule(lr){2-4} \cmidrule(lr){5-7}
         & \textbf{Lane No.} & \textbf{Speed Limit} & \textbf{Collision Location} & \textbf{Driving Actions} & \textbf{Crash Type} & \textbf{Participant No.} \\
        \midrule
        SoVAR~\cite{guo2024sovar} & 93.33 & \textbf{100.00} & 96.00 & 76.33 & \textbf{90.67} & \textbf{100.00} \\
        SoVAR\_N~\cite{guo2024sovar} & 76.00 & 91.33 & 95.33 & 55.00 & 90.00 & \textbf{100.00} \\
        AC3R~\cite{huynh2019ac3r} & 70.67 & 82.67 & 78.67 & 43.67 & 11.67 & 94.67 \\
        \midrule
        \rowcolor{gray!20}\textbf{AccidentSim} & \textbf{98.00} & \textbf{100.00} & \textbf{98.00} & \textbf{93.00} & 88.00 & \textbf{100.00} \\
        \bottomrule
    \end{tabular}}
\end{table*}

\begin{figure}[t]
    \centering
    \includegraphics[width=\columnwidth]{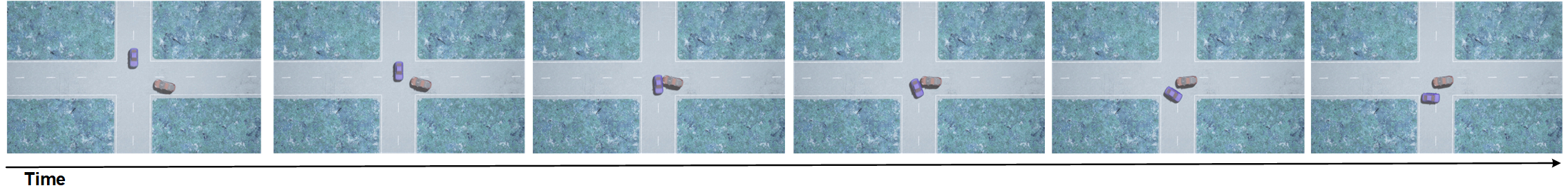}
    \caption{Examples of reconstructed collision scenarios in CARLA. Extracted road and vehicle attributes initialize the scene, while collision sensors record the impact state and post-collision trajectory.}
    \label{fig:carla}
\end{figure}

We next measure Scenario Reconstruction Rate (SRR), following SoVAR~\cite{guo2024sovar}, to determine whether the extracted attributes can be instantiated as valid collision trajectories. AccidentSim improves SRR on intersections, T-junctions, and straight roads (Table~\ref{tab:srr}), with the largest gain of 8.2 percentage points on straight roads. Figure~\ref{fig:carla} illustrates the corresponding CARLA reconstructions used to collect collision-sensor states and physics-constrained post-impact trajectories.
\begin{table}[t]
    \centering
    \caption{Scenario Reconstruction Rate (SRR, \%) across road types.}
    \label{tab:srr}
    \resizebox{\columnwidth}{!}{
    \begin{tabular}{l c c c}
        \toprule
        \textbf{Method} & \textbf{Intersection} & \textbf{T-junction} & \textbf{Straight Road} \\
        \midrule
        SoVAR~\cite{guo2024sovar} & 93.3 & 72.7 & 82.0 \\
        \rowcolor{gray!20}\textbf{AccidentSim} & \textbf{95.0} & \textbf{76.3} & \textbf{90.2} \\
        \bottomrule
    \end{tabular}}
\end{table}
\begin{figure*}[!htb]
    \centering
    \includegraphics[width=\textwidth]
    {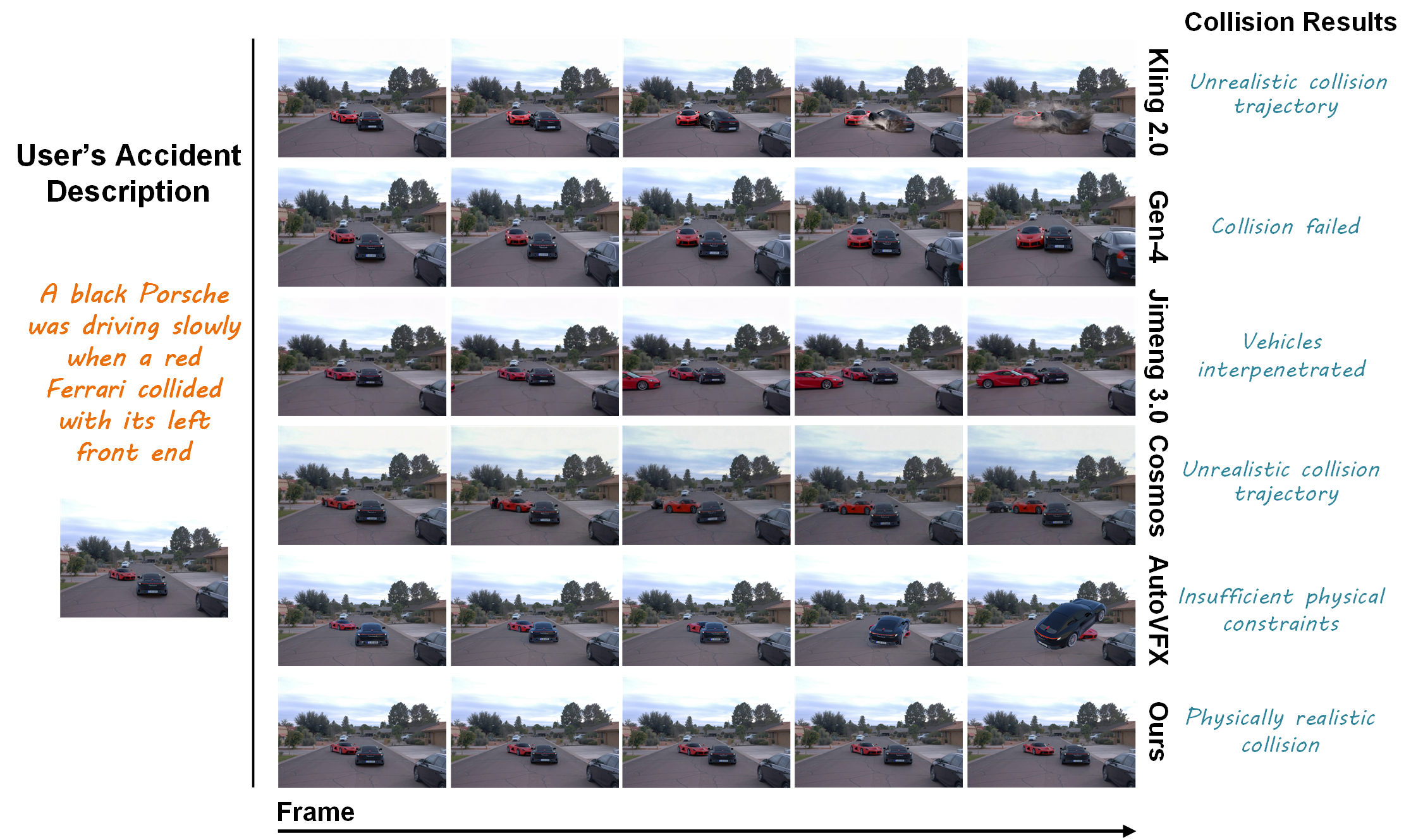}
    \captionsetup{width=\textwidth}  
    \caption{Qualitative comparison under the same user-provided accident description and driving scene. Vehicles involved in the collision are boxed; the final column summarizes the observed collision outcome for each method.}
    \label{fig:collision_comparison}
\end{figure*}
\begin{table*}[t]
\caption{Post-collision trajectory and physical-consistency errors for AccidentSim and AutoVFX across four accident scenarios. AccidentSim attains lower average L2 error in three of the four scenarios.}
\centering
\resizebox{0.7\textwidth}{!}{
\setlength{\tabcolsep}{2.7mm} 
\begin{tabular}{l l c c c c c c c c}
\toprule
\multirow{3}{*}{\textbf{Accident Scenarios}} & \multirow{3}{*}{\textbf{Methods}} & \multicolumn{4}{c}{\textbf{Trajectory Prediction Error}} & \multicolumn{4}{c}{\textbf{Physical Consistency Error}} \\ 
\cmidrule(lr){3-6} \cmidrule(lr){7-10}
& &  \multicolumn{4}{c}{\textbf{L2 (m)}} & \multicolumn{2}{c}{\textbf{Impulse}} & \multicolumn{2}{c}{\textbf{Momentum}} \\ 
\cmidrule(lr){3-6} \cmidrule(lr){7-8} \cmidrule(lr){9-10}
& & \textbf{1s} & \textbf{2s} & \textbf{3s} & \textbf{Average $\pmb{\downarrow}$} & \textbf{MAE $\pmb{\downarrow}$} & \textbf{RMSE $\pmb{\downarrow}$} & \textbf{MAE $\pmb{\downarrow}$} & \textbf{RMSE $\pmb{\downarrow}$} \\ 
\midrule
 & AutoVFX & 1.21 & 1.90 & 2.42 & 1.84  & 2.89 & 2.91 & 2.77 & 2.65 \\ 
\rowcolor{gray!20}
\cellcolor{white}\multirow{-2}{*}{Straight Obstacle} & \textbf{AccidentSim} & 1.00 & 1.45 & 2.01 & \textbf{1.49} & \textbf{2.02} & \textbf{1.98} & \textbf{2.12} & \textbf{2.21} \\ 
\midrule
 & AutoVFX & 1.02 & 1.43 & 1.82 & \textbf{1.42} & \textbf{1.93} & \textbf{1.87} & \textbf{1.76} & \textbf{1.84} \\ 
\rowcolor{gray!20}
\cellcolor{white}\multirow{-2}{*}{Unprotected Left-Turn} & \textbf{AccidentSim} & 1.22 & 1.67 & 2.31 & 1.73 & 2.31 & 2.40 & 2.87 & 2.54 \\ 
\midrule
 & AutoVFX & 1.04 & 2.32 & 2.76 & 2.04 & 3.10 & 3.02 & 2.87 & 2.71  \\ 
\rowcolor{gray!20}
\cellcolor{white}\multirow{-2}{*}{Right Turn} & \textbf{AccidentSim} & 0.45 & 0.83 & 1.54 & \textbf{0.94} & \textbf{0.89} & \textbf{0.94} & \textbf{0.78} & \textbf{0.87} \\ 
\midrule
 & AutoVFX & 1.45 & 1.98 & 2.58 & 2.00  & 3.21 & 3.18 & 2.89 & 2.92 \\ 
\rowcolor{gray!20}
\cellcolor{white}\multirow{-2}{*}{Lane Changing} & \textbf{AccidentSim} & 0.67 & 1.21 & 1.95 & \textbf{1.28}   & \textbf{1.37} & \textbf{1.48} & \textbf{1.54} & \textbf{1.64}  \\ 
\bottomrule
\end{tabular}
}
\label{tab:l2}  
\end{table*}
\begin{table*}[!htb]
\centering
\caption{Human study results on the language-conditioned simulation.}
\label{tab:human_study}
\resizebox{0.7\textwidth}{!}{
\begin{tabular}{l c c c c} 
\toprule
\multirow{2}{*}{\textbf{Methods}} & \multicolumn{2}{c}{\textbf{Crash Report}} & \multicolumn{2}{c}{\textbf{Attribute Description}} \\
\cmidrule(lr){2-3} \cmidrule(lr){4-5} 
& \textbf{Ours Preferred (\%)} $\uparrow$ & \textbf{Score (1--5)} $\uparrow$ & \textbf{Ours Preferred (\%)} $\uparrow$ & \textbf{Score (1--5)} $\uparrow$ \\
\midrule
TrafficGen \cite{feng2022trafficgen} & 95.68 & 1.49  & 94.51 & 2.63 \\
MotionCLIP \cite{tevet2022motionclip}& 96.49 & 1.71 & 95.71 & 2.20 \\
LCTGen \cite{tan2023language} & 89.53 & 3.93 & 87.29 & 4.10 \\
\midrule
\rowcolor{gray!20} 
\textbf{AccidentSim} & \textbf{--} & \textbf{4.49} & \textbf{--} & \textbf{4.46} \\
\bottomrule
\end{tabular}
}
\end{table*}
\begin{table*}[!htb]
    \caption{
    Quality assessment of videos generated by AccidentSim and the compared methods using VBench~\cite{huang2024vbench}, human evaluation~\cite{chen2025physgen3d}, and GPT-4o evaluation~\cite{chen2025physgen3d}. The best value is in bold and the second-best is underlined.}
    \centering
    \resizebox{0.8\textwidth}{!}{
    \begin{tabular}{l c c c c c c c c} 
        \toprule
        \multirow{2}{*}{\textbf{Methods}} & \multicolumn{2}{c}{\textbf{VBench Evaluation \cite{huang2024vbench}}} & \multicolumn{3}{c}{\textbf{Human Evaluation \cite{chen2025physgen3d}}} & \multicolumn{3}{c}{\textbf{GPT-4o Evaluation \cite{chen2025physgen3d}}} \\
        \cmidrule(lr){2-3} \cmidrule(lr){4-6} \cmidrule(lr){7-9} 
         & \textbf{Motion Smoothness $\pmb{\uparrow}$} & \textbf{Imaging Quality $\pmb{\uparrow}$} &\textbf{PhysReal $\pmb{\uparrow}$} & \textbf{PhotoReal $\pmb{\uparrow}$} & \textbf{Align $\pmb{\uparrow}$} & \textbf{PhysReal $\pmb{\uparrow}$} & \textbf{PhotoReal $\pmb{\uparrow}$} & \textbf{Align $\pmb{\uparrow}$} \\
        \midrule
        Kling 2.0 \cite{klingai2025} & 0.991 & 0.697 & 2.350 &  3.186 & 2.872 & 0.542 & 0.758 & 0.571 \\
        Gen-4 \cite{Runway2025} & 0.993 & 0.686 &  2.456  & 3.497  & 2.403 & 0.614 & \textbf{0.891} & 0.492 \\
        Jimeng 3.0 \cite{Jimengai2025} & \underline{0.996} & \textbf{0.727} & \underline{3.011} & \textbf{3.703 } & 3.219 & \underline{0.724} & 0.855 & \underline{0.724} \\
        Cosmos \cite{agarwal2025cosmos} & 0.985 & 0.637 &  2.578 & 3.142 & 2.894 & 0.451 & 0.703 & 0.510 \\
        AutoVFX \cite{hsu2024autovfx} & 0.990 & 0.701 & 2.983 & 3.286 & \underline{3.450} & 0.521 & 0.768 & 0.682 \\
        \midrule 
        \rowcolor{gray!20} 
        \textbf{AccidentSim} & \textbf{0.997} & \underline{0.718} & \textbf{3.906} &  \underline{3.622} &  \textbf{4.186} & \textbf{0.839} & \underline{0.873} & \textbf{0.910} \\
        \bottomrule
    \end{tabular}
    }
    \label{tab:comprehensive_eval}
\end{table*}
\begin{table*}[t]
\caption{Collision rates for policies trained with scenarios from AccidentSim and the baseline generators. Lower is better; the best value is in bold and the second-best is underlined.}
\centering
\resizebox{\textwidth}{!}{
\begin{tabular}{l c c c c c c c c c c c c c} 
\toprule
\multirow{4}{*}{\textbf{Methods}} & \multicolumn{4}{c}{\textbf{Dual-Vehicle Scenarios}} & \multicolumn{4}{c}{\textbf{Triple-Vehicle Scenarios}} & \multicolumn{4}{c}{\textbf{Multi-Vehicle Scenarios}} & \multirow{2}{*}{\textbf{Avg.}} \\ 
\cmidrule(lr){2-5} \cmidrule(lr){6-9} \cmidrule(lr){10-13}
& \shortstack{\textbf{Straight}\\\textbf{Obstacle}} & \shortstack{\textbf{Unprotected}\\\textbf{Left-Turn}} & \shortstack{\textbf{Right-}\\\textbf{Turn}} & \shortstack{\textbf{Lane}\\\textbf{Changing}} 
& \shortstack{\textbf{Straight}\\\textbf{Obstacle}} & \shortstack{\textbf{Unprotected}\\\textbf{Left-Turn}} & \shortstack{\textbf{Right-}\\\textbf{Turn}} & \shortstack{\textbf{Lane}\\\textbf{Changing}} 
& \shortstack{\textbf{Straight}\\\textbf{Obstacle}} & \shortstack{\textbf{Unprotected}\\\textbf{Left-Turn}} & \shortstack{\textbf{Right-}\\\textbf{Turn}} & \shortstack{\textbf{Lane}\\\textbf{Changing}} & \\
\midrule
Pre Fine-tuning 
& 0.48 & 0.67 & 0.46 & 0.58
& 0.52 & 0.59 & 0.47 & 0.63
& 0.43 & 0.60 & 0.51 & 0.68
& 0.552 \\ 
Learning-to-collide \cite{Ding2020LearningTC}
& 0.12 & \textbf{0.00} & 0.37 & 0.51
& 0.22 & 0.13 & 0.40 & 0.52
& 0.28 & 0.25 & 0.39 & 0.46
& 0.304 \\ 
AdvSim \cite{Wang2021AdvSimGS} 
& 0.23 & \underline{0.05} & 0.41 & 0.53
& 0.31 & \underline{0.10} & 0.38 & 0.45
& 0.28 & 0.17 & 0.35 & 0.58
& 0.320 \\ 
Carla Scenario Generator \cite{contributors2019carla} 
& 0.22 & 0.22 & \underline{0.19} & 0.39
& 0.17 & 0.26 & 0.17 & 0.48
& 0.12 & 0.29 & \underline{0.10} & 0.48
& 0.258 \\ 
Adversarial Trajectory Optimization \cite{zhang2022adversarial}
& 0.14 & \textbf{0.00} & 0.23 & 0.30 
& 0.21 & \textbf{0.04} & 0.28 & \underline{0.21}
& 0.23 & \textbf{0.02} & 0.31 & \underline{0.18}
& 0.179 \\ 
ChatScene \cite{zhang2024chatscene}
& \textbf{0.03} & 0.10 & \textbf{0.01} & \underline{0.11}
& \underline{0.09} & 0.21 & \textbf{0.06} & 0.24
& \underline{0.05} & 0.16 & \textbf{0.09} & 0.19
& \underline{0.112} \\
\midrule 
\rowcolor{gray!20} 
\textbf{AccidentSim} 
&  \underline{0.05} & 0.08 & \textbf{0.01} & \textbf{0.09}
& \textbf{0.07} & 0.11 & \underline{0.13} & \textbf{0.12}
& \textbf{0.02} & \underline{0.07} & \underline{0.10} & \textbf{0.03}
& \textbf{0.073} \\ 
\bottomrule
\end{tabular}
}
\label{tab:collision}  
\end{table*}
\begin{figure*}[hbt!]
\centering
\includegraphics[width=0.8\textwidth]
{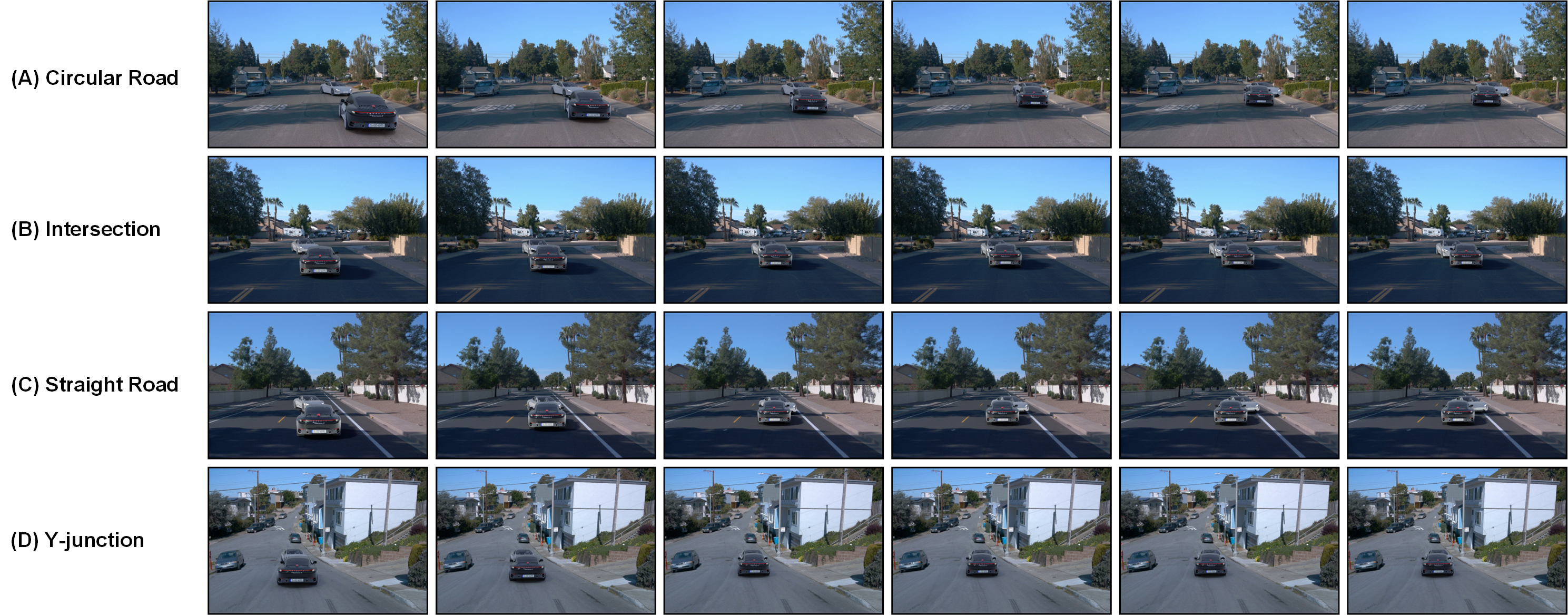}
\caption{AccidentSim-generated collision videos across four road types and multiple viewpoints. Each row shows temporally ordered frames from one generated sequence.}
\label{fig:scene_result}  
\end{figure*}
\subsection{Trajectory Generation Quality and Physical Consistency}\label{sec:traj_quality}
\textit{1) Physical Consistency and Accuracy:} We evaluate the physical plausibility of AccidentLLM's post-collision trajectories against CARLA-simulated ground truth. \textbf{Trajectory Prediction Error} is measured via L2 distance, and \textbf{Physical Consistency Error} is assessed using the MAE and RMSE of impulse and momentum changes \cite{zhou2008collision}. We select AutoVFX \cite{hsu2024autovfx}, a physics-based simulation method, as our primary baseline. As shown in Table~\ref{tab:l2}, AccidentSim obtains lower average L2 error in three of the four scenarios and lower impulse and momentum errors in the same three scenarios; AutoVFX performs better on the unprotected left-turn case. Figure~\ref{fig:collision_comparison} further visualizes the different collision trajectories produced by the compared methods.

\textit{2) Text-to-Trajectory Semantic Consistency:} Following the LCTGen \cite{tan2023language} protocol, we conducted human A/B tests to measure Preference (Ours Preferred \%) and Alignment (1--5 scores) on Crash Report and Attribute Description datasets. Table~\ref{tab:human_study} shows that AccidentSim receives higher preference and alignment scores than the compared baselines. These results support the use of simulator-calibrated data for mapping textual narratives to kinematically consistent trajectories.
\subsection{Trajectory Fidelity and Prediction Utility}\label{sec:prediction_enhancement}
We complement the collision-specific errors in Table~\ref{tab:l2} with distribution-level trajectory metrics following LCTGen~\cite{tan2023language}. Maximum Mean Discrepancy (MMD) measures alignment of scene initialization attributes, including position, heading, speed, and vehicle size, while mean Average Displacement Error (mADE) and mean Final Displacement Error (mFDE) measure generated motion. The scenario reconstruction score (SCR) follows LCTGen and is distinct from the percentage-based SRR in Table~\ref{tab:srr}. Table~\ref{tab:results} shows that AccidentSim substantially reduces initialization mismatch and mADE. Its mFDE is close to the best baseline, while its higher SCR indicates that the generated trajectories more often instantiate valid scenes.
\begin{table*}[t]
\centering
\caption{Distribution-level fidelity of generated traffic scenarios. Lower is better for MMD, mADE, and mFDE; higher is better for the scenario reconstruction score (SCR).}
\label{tab:results}
\resizebox{0.72\textwidth}{!}{
\begin{tabular}{l c c c c c c c}
\toprule
\multirow{2}{*}{\textbf{Method}} & \multicolumn{4}{c}{\textbf{Initialization MMD $\downarrow$}} & \multicolumn{2}{c}{\textbf{Motion $\downarrow$}} & \multirow{2}{*}{\textbf{SCR $\uparrow$}} \\
\cmidrule(lr){2-5} \cmidrule(lr){6-7}
 & \textbf{Position} & \textbf{Heading} & \textbf{Speed} & \textbf{Size} & \textbf{mADE} & \textbf{mFDE} & \\
\midrule
TrafficGen~\cite{feng2022trafficgen} & 0.2002 & 0.1524 & 0.2379 & 0.0951 & 10.448 & 20.646 & 5.690 \\
MotionCLIP~\cite{tevet2022motionclip} & 0.1236 & 0.1446 & 0.1958 & 0.1234 & 6.683 & 13.421 & 8.842 \\
LCTGen~\cite{tan2023language} & 0.0616 & 0.1154 & 0.0719 & 0.1203 & 1.329 & \textbf{2.838} & 6.700 \\
\midrule
\rowcolor{gray!20}\textbf{AccidentSim} & \textbf{0.0325} & \textbf{0.0651} & \textbf{0.0324} & \textbf{0.0870} & \textbf{0.925} & 2.972 & \textbf{8.925} \\
\bottomrule
\end{tabular}}
\end{table*}

To test whether the generated trajectories provide useful supervision beyond the AccidentSim domain, we follow the cross-dataset protocol of Traj-LLM~\cite{yang2025trajectory}. Synthetic AccidentSim trajectories augment the training data of MTR~\cite{shi2022motion} and HDGT~\cite{jia2023hdgt}; evaluation is performed on both WOMD validation data and held-out AccidentSim trajectories. As shown in Table~\ref{tab:accident_sim_results}, joint training consistently improves mAP, minADE, and miss rate over either single-source training regime. This result indicates that the generated safety-critical trajectories add complementary supervision without sacrificing performance on normal-driving data.
\begin{table*}[t]
\centering
\caption{Downstream trajectory-prediction results. Each entry reports mAP $\uparrow$ / minADE $\downarrow$ / miss rate $\downarrow$.}
\label{tab:accident_sim_results}
\resizebox{0.82\textwidth}{!}{
\begin{tabular}{c c c c c c}
\toprule
\multicolumn{2}{c}{\textbf{Training data}} & \multicolumn{2}{c}{\textbf{WOMD validation}} & \multicolumn{2}{c}{\textbf{AccidentSim validation}} \\
\cmidrule(lr){1-2} \cmidrule(lr){3-4} \cmidrule(lr){5-6}
\textbf{WOMD} & \textbf{AccidentSim} & \textbf{MTR} & \textbf{HDGT} & \textbf{MTR} & \textbf{HDGT} \\
\midrule
\checkmark & & 0.389/0.728/0.116 & 0.326/0.572/0.216 & 0.303/0.872/0.249 & 0.272/1.321/0.408 \\
 & \checkmark & 0.143/1.921/0.342 & 0.201/2.329/0.310 & 0.385/0.734/0.136 & 0.216/0.724/0.338 \\
\midrule
\rowcolor{gray!20}\checkmark & \checkmark & \textbf{0.452/0.692/0.102} & \textbf{0.332/0.546/0.189} & \textbf{0.418/0.703/0.129} & \textbf{0.298/0.684/0.301} \\
\bottomrule
\end{tabular}}
\end{table*}
\subsection{Quality Evaluation for Generated Video}
\label{sec:video_quality}

This experiment evaluates the visual fidelity and physical realism of generated vehicle collisions. Since conventional metrics (e.g., PSNR, FID, FVD) require ground-truth data, we adopt the PhysGen3D \cite{chen2025physgen3d} protocol for reference-free evaluation. First, we use VBench \cite{huang2024vbench} to assess motion smoothness and imaging quality. Second, we employ human and GPT-4o \cite{GPT-4o} evaluations to measure three key metrics: Physical Realism (PhysReal), Photorealism (Photoreal), and Semantic Consistency (Align). 

Specifically, \textbf{PhysReal} measures adherence to physical laws, capturing dynamics like gravity and momentum; \textbf{Photoreal} assesses overall visual fidelity, focusing on the absence of artifacts and the accurate depiction of lighting and shadows; and \textbf{Align} evaluates how accurately the generated video reflects the semantics of the user-provided accident description.

Detailed human- and GPT-4o-evaluation protocols, the complete human-rating distributions, and the exact prompt used for GPT-4o evaluation are provided in the supplementary material.

We compare AccidentSim against five state-of-the-art video generation models: Kling 2.0 \cite{klingai2025}, Gen-4 \cite{Runway2025}, Jimeng 3.0 \cite{Jimengai2025}, Cosmos \cite{agarwal2025cosmos}, and AutoVFX \cite{hsu2024autovfx}. For a fair comparison, all models receive identical accident descriptions and background scenes.

\begin{figure}[H]
    \centering
    \includegraphics[width=\columnwidth]{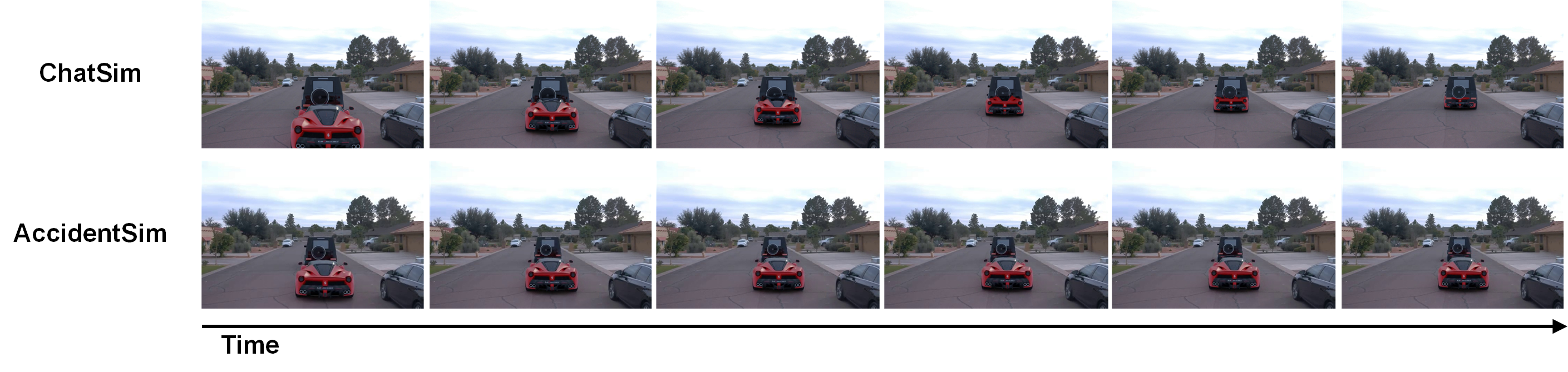}
    \caption{Frame-by-frame comparison of collision dynamics. ChatSim exhibits vehicle-interpenetration artifacts, whereas AccidentSim maintains distinct vehicle boundaries during collision.}
    \label{fig:frame_by_frame_collision}
\end{figure}

\textit{Additional Frame-by-Frame Collision Analysis:} Figure~\ref{fig:frame_by_frame_collision} compares temporally ordered collision frames generated by AccidentSim and ChatSim~\cite{wei2024editable}. ChatSim exhibits vehicle-interpenetration artifacts during contact, whereas AccidentSim maintains distinct vehicle boundaries and produces a physically plausible collision response.

As shown in Table~\ref{tab:comprehensive_eval}, AccidentSim achieves the best motion smoothness, physical realism, and semantic consistency among the compared methods, while ranking second in VBench imaging quality and GPT-4o photorealism. Qualitative comparisons in Figure~\ref{fig:collision_comparison} 
and diverse scenario results in Figure~\ref{fig:scene_result} further illustrate the generated collision behavior across methods and scenarios.

\subsection{Collision Reduction by AccidentSim}
\label{sec:collision_reduction}
We evaluate whether AccidentSim-generated collision scenarios improve the safety of an autonomous-driving policy. Experiments use SafeBench~\cite{xu2022safebench} on an NVIDIA A40 GPU. We transfer complex NHTSA accident descriptions to road scenes derived from the Waymo Open Dataset and express the resulting configurations in Scenic~\cite{fremont2019scenic} for execution in CARLA. The benchmark contains 4,000 training scenarios and 1,000 testing scenarios. Following the taxonomy of ChatScene~\cite{zhang2024chatscene}, we cover dual-vehicle scenarios and extend the evaluation to triple- and multi-vehicle collisions.

We train the vehicle-control policy with Soft Actor-Critic (SAC)~\cite{haarnoja2018soft} for 500 epochs using a learning rate of $10^{-4}$ and evaluate it every 50 epochs. ``Pre Fine-tuning'' denotes the SAC policy before augmentation with AccidentSim scenarios. We compare against Learning-to-Collide~\cite{Ding2020LearningTC}, AdvSim~\cite{Wang2021AdvSimGS}, CARLA Scenario Generator~\cite{contributors2019carla}, Adversarial Trajectory Optimization~\cite{zhang2022adversarial}, and ChatScene~\cite{zhang2024chatscene}. As shown in Table~\ref{tab:collision}, AccidentSim reduces the average collision rate to 0.073, a 34.8\% relative reduction from ChatScene's 0.112 and the best average result among the compared methods. These findings demonstrate the practical value of physically grounded collision scenarios for safety-oriented policy training.

\subsection{Limitations and Future Work}\label{sec:limitations}
AccidentSim focuses on rigid-body vehicle motion and does not yet model detailed deformation, debris, or occupant-level effects. Its fidelity also depends on the accuracy of attributes extracted from accident narratives; missing or ambiguous speeds, headings, and road conditions can lead to multiple physically plausible reconstructions. Finally, the current evaluation is conducted in CARLA and on selected road and collision configurations, so performance may change under other simulators, vehicle dynamics, weather conditions, and sensor domains. Future work will incorporate deformable collision models, uncertainty-aware attribute extraction, and cross-simulator validation.
\section{Conclusion}
In this paper, we introduce AccidentSim, a framework for generating vehicle collision videos with physically grounded collision trajectories. AccidentSim extracts physical cues and contextual information from real-world accident reports, uses CARLA to construct a collision-trajectory dataset, and fine-tunes AccidentLLM for post-collision trajectory prediction. A pre-collision planner and the predicted post-collision motion jointly form complete trajectories for video rendering. Across the evaluated scenarios, AccidentSim achieves the best overall physical-realism and semantic-consistency results among the compared methods, improves downstream trajectory prediction when used for data augmentation, and reduces the average collision rate of the evaluated driving policy. Future work will incorporate vehicle deformation, uncertainty-aware report extraction, and cross-simulator evaluation.

\bibliographystyle{IEEEtran}
\bibliography{references}
\begin{IEEEbiography}[{\includegraphics[width=1in,height=1.25in,clip,keepaspectratio]{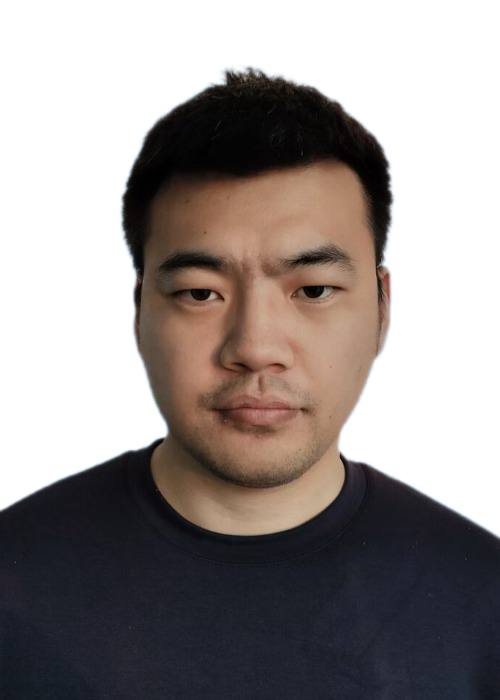}}]{Xiangwen Zhang}
is currently pursuing the master's degree with the School of Computer and Artificial Intelligence, Beijing Technology and Business University, Beijing, China. His current research interests include autonomous driving, trajectory prediction, and generative models.
\end{IEEEbiography}
\begin{IEEEbiography}[{\includegraphics[width=1in,height=1.25in,clip,keepaspectratio]{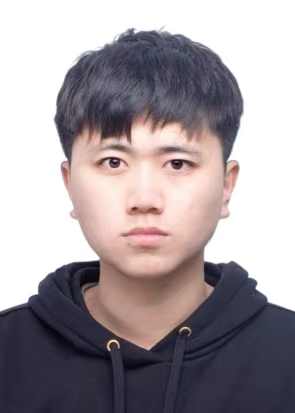}}]{Qian Zhang} is currently pursuing the master's degree with Beijing Technology and Business University, Beijing, China, under the supervision of Prof. Xiaoming Chen. His research interests include 3D Gaussian splatting and artificial intelligence.\end{IEEEbiography}
\begin{IEEEbiography}[{\includegraphics[width=1in,height=1.25in,clip,keepaspectratio]{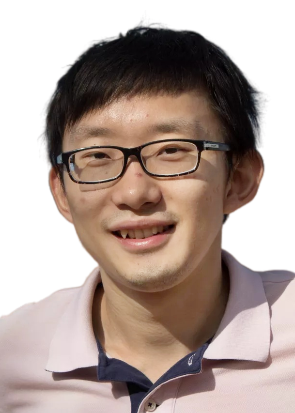}}]{Longfei Han} received the Ph.D. degree from Beijing
Institute of Technology. He was a Ph.D. Visiting
Student at Carnegie Mellon University. He is an
Associate Professor at the School of Computer
and Artificial Intelligence, Beijing Technology and
Business University. After graduation, he worked as a
Senior Engineer at Tencent, focusing on
computational advertising. His current research interests include
large-scale pretrained frameworks, lightweight
neural networks, and multimodal learning.\end{IEEEbiography}
\begin{IEEEbiography}[{\includegraphics[width=1in,height=1.25in,clip,keepaspectratio]{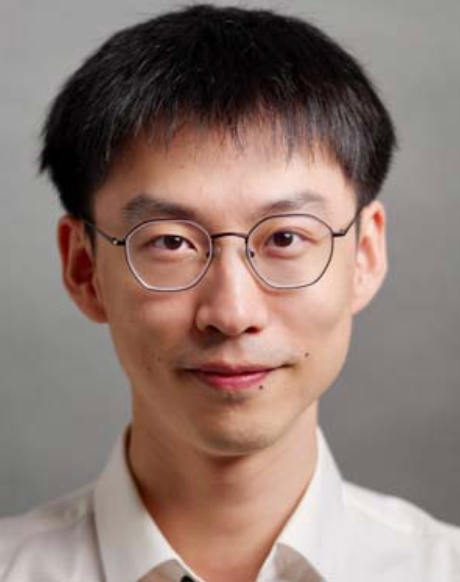}}]{Qiang Qu} received the BE (with Hons.) and BCom
degrees in software engineering and commerce, and
the PhD degree in computer science from the University of Sydney, Camperdown, NSW, Australia.
His research focuses on automated perceptual quality
assessment for AI-generated immersive media, including light field images, 3D reconstructions, and
VR/AR/MR content. He also conducts research in
neuromorphic vision (event cameras), developing efficient processing algorithms for bio-inspired visual
sensors. His work has been authored or coauthored in
leading journals and conferences, such as \textit{IEEE Transactions on Pattern Analysis
and Machine Intelligence} (TPAMI), \textit{IEEE Transactions on Visualization and
Computer Graphics} (TVCG), \textit{IEEE Transactions on Image Processing} (TIP),
\textit{IEEE Transactions on Broadcasting}, IEEE VR, and AAAI. \end{IEEEbiography}
\begin{IEEEbiography}[{\includegraphics[width=1in,height=1.25in,clip,keepaspectratio]{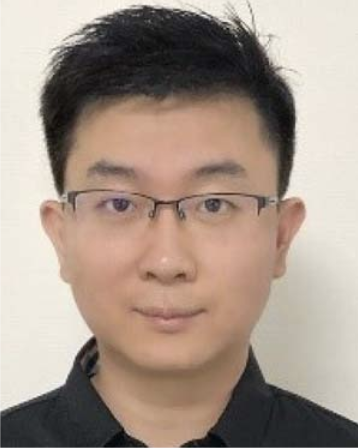}}]{Xiaoming Chen}  received the B.Sc. (with Distinction) degree from the Royal Melbourne Institute of
Technology, Melbourne, VIC, Australia, and the PhD
degree from the University of Sydney, Australia. He
has held positions in both academia and industry,
including roles with the National University of Singapore, Singapore, Nanyang Technological University,
Singapore, CSIRO, Technicolor Research, IBM, Intel
Collaborative Research Institutes, and the University
of Science and Technology of China (USTC), China.
He is currently a Professor with the School of Computer and Artificial Intelligence, Beijing Technology and Business University
(BTBU), Beijing, China. His research interests include immersive media computing, virtual reality, and bio-inspired event processing. His work has appeared
in leading venues, such as \textit{IEEE Transactions on Image Processing}, \textit{IEEE
Transactions on Visualization and Computer Graphics}, \textit{IEEE Transactions on
Circuits and Systems for Video Technology}, IEEE VR, ACM MM, ECCV, and AAAI. \end{IEEEbiography}
\begin{IEEEbiography}[{\includegraphics[width=1in,height=1.25in,clip,keepaspectratio]{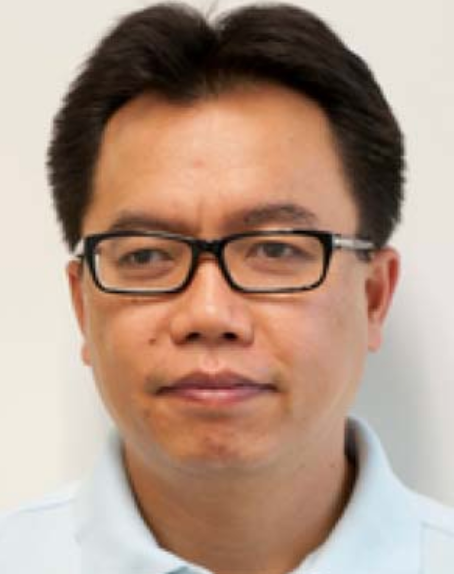}}]{Weidong Cai} is currently an associate professor with the School of Computer Science, Director of the Multimedia Laboratory, and the associate director with the Biomedical and Multimedia Information Technology (BMIT) Research Group, University of
Sydney, Australia. In 2014, he was a lead investigator/visiting professor on medical image analysis and medical computer vision with Harvard Medical School. He has authored or coauthored more than 300 peer-reviewed papers in leading international journals and proceedings of top international conferences. His research interests include computer vision, medical image computing, machine learning, image/video processing, multimedia computing, pattern recognition, computer graphics, bioimage informatics, and computational neuroscience. He is also a senior area editor of \textit{IEEE Transactions on Image Processing}, an associate editor for \textit{Computational Visual Media and Brain Informatics}.\end{IEEEbiography}
\end{document}